\RequirePackage{amsmath} % here as a fix for the \vec warning
\documentclass{llncs}
\usepackage[utf8]{inputenc}
\usepackage{amssymb}
\usepackage{tikz}
\usepackage{multirow}

\graphicspath{{./figures/}}

\begin{document}

\title{Differentiable Genetic Programming}
\titlerunning{dCGP} 
\author{Dario Izzo\inst{1} \and Francesco Biscani\inst{2} \and Alessio Mereta\inst{1}}
\authorrunning{Dario Izzo et al.} % abbreviated author list (for running head)
\institute{Advanced Concepts Team, European Space Agency,  Noordwijk 2201AZ, The Netherlands\\
\email{dario.izzo@esa.int}
%\and
%Advanced Concepts Team, European Space Technology Center,  Noordwijk 2201AZ, %The Netherlands\\
%\email{second.author@gmail.com}
}
%\author{}
%\institute{}

\maketitle              % typeset the title of the contribution

\begin{abstract}
We introduce the use of high order automatic differentiation,
implemented via the algebra of truncated Taylor polynomials, in genetic
programming. Using the Cartesian Genetic Programming encoding we obtain a
high-order Taylor representation of the program output that is then used to
back-propagate errors during learning. The resulting machine learning
framework is called differentiable Cartesian Genetic Programming (dCGP).
In the context of symbolic regression, dCGP offers a new approach
to the long unsolved problem of constant representation in GP
expressions. On several  problems of increasing complexity we
find that dCGP is able to find the exact form of the symbolic
expression as well as the constants values. We also demonstrate the use
of dCGP to solve a large class of differential equations and to find
prime integrals of dynamical systems, presenting, in both cases, results
that confirm the efficacy of our approach.

\keywords{genetic programming, truncated Taylor polynomials, machine learning, symbolic regression, back-propagation}
\end{abstract}
%(posx, posy) with [function text, node id text, radius, input text top, bottom]

\section{Introduction}
Many of the most celebrated techniques in Artificial Intelligence
would not be as successful if, at some level, they were not
exploiting differentiable quantities. The back-propagation algorithm,
at the center of learning in artificial neural networks, leverages on
the first and (sometimes) second order differentials of the error to
update the network weights. Gradient boosting techniques make use of
the negative gradients of a loss function to iteratively improve over
some initial model. More recently, differentiable memory access operations were successfully
implemented \cite{graves1} \cite{graves2} and shown to give rise to new, exciting, neural
architectures. Even in the area of evolutionary computations, also sometimes referred to as
\emph{derivative-free} optimization, having the derivatives of the fitness function is immensely
useful, to the extent that many derivative-free algorithms, in one way or another, seek to approximate
such information (e.g. the covariance matrix in CMA-ES as an approximation of the inverse Hessian
\cite{hansen2006cma}). In the context of Genetic Programming too, previous works made use of the
differential properties of the encoded programs \cite{lipson} \cite{tsoulos} \cite{erc1} \cite{erc2},
showing the potential use of such information, though a systematic use of the  program differentials is still lacking in this field.

In this paper, we introduce the use of high-order automatic differentiation
as a tool to obtain a complete representation of the differential properties
of a program encoded by a genetic programming expression and thus of the model it represents. We make use of
the truncated Taylor polynomial algebra to
compute such information efficiently in one forward pass of the encoded program.
The non trivial implementation of
the necessary algebraic manipulations, as well as the resulting framework,
is offered to the community as an open source project called AuDi \cite{audi} (a C++11
header only library and a python module) allowing the machine learning community
to gain access to a tool we believe can lead to several interesting advances.
We use AuDi to evaluate Cartesian Genetic Programming expressions introducing a number
of applications where the differential information is used to back-propagate the error on
a number of model parameters. The resulting new machine learning tool is called dCGP and
is also released as an open source project \cite{dCGP}. Using dCGP we study in more detail three distinct
application domains where the Taylor expansion of the encoded program is
used during learning: we perform symbolic regression on expressions
including real constants, we search for analytical solutions to
differential equations, we search prime integrals of motion in dynamical
systems. 

The paper is structured as follows: in Section \ref{sec:encoding} we describe the program
encoding used, essentially a weighted version of the standard CGP
encoding. In Section \ref{sec:taylor} we introduce, the
differential algebra of truncated Taylor polynomial used to obtain a
high order automatic differentiation system. Some examples (Section \ref{sec:example}) are also given to help the
reader go through the possibly unfamiliar notation and algebra. In Section \ref{sec:symbolic}, using
dCGP, we propose two different new methods
to perform symbolic regression on expressions that include real constants. In Section \ref{sec:de} we
use dCGP
to find the analytical solution to differential equations. In the following Section
\ref{sec:primeintegrals}, we propose a method to systematically search for prime integrals of motions in dynamical systems.

\section{Program encoding}
\label{sec:encoding}
\def\cgpnode at (#1,#2) with [#3,#4,#5, #6, #7] {
\draw (#1,#2) circle (#5);

\node[] at (#1, #2) {#3};
\node[align=left, text width=#5*4cm] at (#1 + #5 * 3.25 , #2+#5*0.3) {#4};
\node[align=right, text width=#5*2cm] at (#1 - #5 * 2.4 , #2+1.1*#5) {#6};
\node[align=right, text width=#5*2cm] at (#1 - #5 * 2.4 , #2-1.1*#5) {#7};

\draw (#1 + #5, #2) -- (#1 + #5*1.7, #2);

\draw [loosely dotted] (#1 - #5*1.2, #2 + 0.1*#5) -- (#1 - #5*1.2, #2 - 0.7*#5); 

\begin{scope}[rotate around={180-45:(#1,#2)}]
  \draw (#1 + #5, #2) -- (#1 + #5*1.5, #2);
\end{scope}

\begin{scope}[rotate around={155:(#1,#2)}]
  \draw (#1 + #5, #2) -- (#1 + #5*1.3, #2);
\end{scope}

\begin{scope}[rotate around={180 + 45:(#1,#2)}]
  \draw (#1 + #5, #2) -- (#1 + #5*1.5, #2);
\end{scope}
} % END CGPNODE

\def\inputnode at (#1,#2) with [#3, #4] {
\draw (#1,#2) circle (#4);
\draw (#1 + #4, #2) -- (#1 + #4*1.7, #2);
\node[align=left, text width=#4*0.5cm] at (#1 + #4*1.5, #2 + #4*0.5) {#3};
} % ENDINPUT

\def\outputnode at (#1,#2) with [#3, #4] {
\draw (#1,#2) circle (#4);
\draw (#1 - #4, #2) -- (#1 - #4*1.7, #2);
\node[align=right, text width=#4*0.5cm] at (#1 - #4*2.5, #2 + #4*0.5) {#3};
} % ENDOUTPUT

\begin{figure}
\center
\resizebox{1.0\textwidth}{!}{%
\begin{tikzpicture}
\cgpnode at (0,0) with [$F_1$, $n+1$,0.8, $C_{1,1}$, $C_{1,a}$]
\cgpnode at (0,-2.5) with [$F_{2}$, $n+2$,0.8, $C_{2,1}$, $C_{2,a}$]
\cgpnode at (0,-6) with [$F_{r}$, $n+r$,0.8, $C_{r,1}$, $C_{r,a}$]
\draw [loosely dotted] (0,-3.7) -- (0,-4.9);

\cgpnode at (4.5,0) with [$F_{r+1}$, $n+r+1$,0.8, $C_{r+1, 1}$, $C_{r+1, a}$]
\cgpnode at (4.5,-2.5) with [$F_{r+2}$, $n+r+2$,0.8, $C_{r+2, 1}$, $C_{r+2, a}$]
\cgpnode at (4.5,-6) with [$F_{2r}$, $n+r+r$,0.8, $C_{2r, 1}$, $C_{2r, a}$]
\draw [loosely dotted] (4.5,-3.7) -- (4.5,-4.9);

\cgpnode at (11,0) with [$F_{cr+1}$, $n+cr+1$,0.8, $C_{cr+1, 1}$, $C_{cr+1, a}$]
\cgpnode at (11,-2.5) with [$F_{cr+2}$, $n+cr+2$,0.8, $C_{cr+2, 1}$, $C_{cr+2, a}$]
\cgpnode at (11,-6) with [$F_{r(c+1)}$, $n+cr+r$,0.8, $C_{r(c+1), 1}$, $C_{r(c+1)-1, a}$]
\draw [loosely dotted] (11,-3.7) -- (11,-4.9);

\inputnode at (-5,-0.5) with [$1$, 0.5]
\inputnode at (-5,-2.5) with [$2$, 0.5]
\inputnode at (-5,-5.5) with [${n}$, 0.5]
\draw [loosely dotted] (-5,-3.5) -- (-5,-4.5);

\outputnode at (16,-0.5) with [$o_1$, 0.5]
\outputnode at (16,-2.5) with [$o_2$, 0.5]
\outputnode at (16,-5.5) with [$o_{m}$, 0.5]
\draw [loosely dotted] (16,-3.5) -- (16,-4.5);

\draw [loosely dotted] (7,0) -- (9,0);
\draw [loosely dotted] (7,-2.5) -- (9,-2.5);
\draw [loosely dotted] (7,-6) -- (9,-6);

\node[align=center, text width=20cm] at (5.5, -8) {$F_1, C_{1,1} .. C_{1,a}, F_2, C_{2,1} .. C_{2,a}, ..., o_1 .. o_m, w_{1,1} .. w_{1,a}, ..., w_{2,1} .. w_{2,a}$};

\end{tikzpicture}
}
\caption{The general form of a CGP as described in \cite{millerbook}. In our version weights are also assigned to the connections $C_{ij}$ so that the program is defined by some integer values (the connections, the functions and the outputs) and by some floating point values (the connection weights)\label{fig:cgp}}
\end{figure}
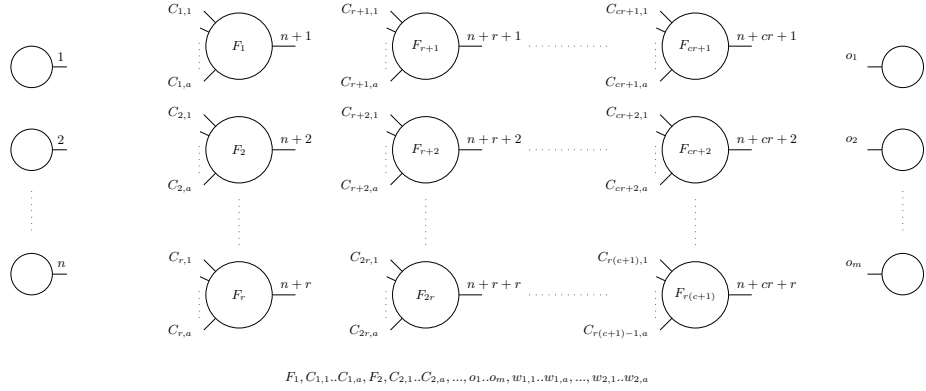
To represent our functional programs we use the Cartesian Genetic Programming framework
\cite{millerbook}. CGP is a form of Genetic Programming in which computations are
organized using a matrix of nodes as depicted in Figure \ref{fig:cgp}. In the original
formulation, nodes are arranged in $r$ rows and $c$ columns, and each is defined by a function
$F_i$ with arity $a$ and by its connections $C_{ij}$, $j = 1 .. a$ indicating the nodes
whose outputs to use to compute the node output via $F_i$. Connections can only point
to nodes in previous columns and a maximum of $l$ columns back (levels back). Input and
output nodes are also present. By evaluating all nodes in cascade starting from the
$N_{in}$ input nodes, the $N_{out}$ outputs nodes are computed. They represent a
complex combination of the inputs able to easily represent computer programs, digital
circuits, mathematical relations and, in general, computational structures. Depending
on the various $C_{ij}$, not all nodes affects the output so that only the active nodes
need to be actually computed. The connections $C_{ij}$ are also here associated to
multiplicative factors (weights) $w_{i,j}$. In this way, a CGP is allowed to represent
highly successful models such as feed forward artificial neural networks as suggested
in \cite{cgpannkahn}, \cite{cgpannturner}.

\section{The algebra of truncated polynomials}
\label{sec:taylor}
Consider the set $\mathcal P_n$ of all polynomials of order $\le n$ with coefficients in $\mathbb R$. Under the truncated polynomial multiplication $\mathcal P_n$ becomes a field (\emph{i.e.}, a ring whose nonzero elements form an abelian group under such multiplication). The meaning of this last statement is, essentially, that we may operate in $\mathcal P_n$ using four arithmetic operations $+,-,\cdot,/$ as we normally do in more familiar fields such as $\mathbb R$ or $\mathbb C$. In order to formally define the division operator, we indicate the generic element of $\mathcal P_n$ as $p = p_0 + \hat p$ so that the constant and the non-constant part of the polynomial are clearly separated. The multiplicative inverse of $p \in \mathcal P_n$ is then defined as:
\begin{equation}
\label{eq:div}
p^{-1} =  1 / p =  \frac 1{p_0} \left(1 +\sum_{k=1}^m (-1)^k (\hat p / p_0)^k\right)
\end{equation}
As an example, to compute, in $\mathcal P_2$, the inverse of $p = 1 + x - y^2$ we write $p_0 = 1$ and $\hat p = x - y^2$. Applying the definition we get $p^{-1} = 1 - \hat p +\hat p^2 = (1 - (x - y^2) + x^2)$. It is then trivial to verify that $p \cdot p^{-1}=1$.

\subsection{The link to Taylor polynomials}

We make use of the multi-index notation according to which $\alpha = (\alpha_1, ..,\alpha_n)$ and $\mathbf x = (x_1, .., x_n)$ are n-tuples and the Taylor expansion around the point $\mathbf a$ to order $m$ of a multivariate function $f$ of $\mathbf x$ is written as:
\begin{equation}
\label{eq:taylor}
T_f(\mathbf x) = \sum_{|\alpha| = 0}^m  \frac{(\mathbf x-\mathbf a)^\alpha}{\alpha!}(\partial^\alpha f)(\mathbf a)
\end{equation}
where:
$$
\partial^\alpha f = \frac{\partial^{|\alpha|} f}{\partial^{\alpha_1} x_1\partial^{\alpha_2} x_2\dots\partial^{\alpha_n} x_n}
$$
$$
\alpha ! = \prod_{j=1}^n \alpha_j!
$$
and 
$$
|\alpha| = \sum_{j=1}^n \alpha_j
$$
The summation $ \sum_{|\alpha| = 0}^n$ must then be taken over all possible combinations of $\alpha_j \in N$ such that $\sum_{j=1}^n \alpha_j = |\alpha|$. 
The expression in Eq.(\ref{eq:taylor}), \emph{i.e.} the Taylor expansion truncated at order $m$ of a generic function $f$, is a polynomial $\in \mathcal P_{n}$ in the $m$ variables $\mathbf{dx} = \mathbf x-\mathbf a$. It follows from Eq.(\ref{eq:taylor}) that a Taylor polynomial contains the information on all the derivatives of $f$.

The remarkable thing about the field $\mathcal P_n$ is that its arithmetic represents also Taylor polynomials, in the sense that if $T_f, T_g \in \mathcal P_{n}$ are truncated Taylor expansions of two functions $f, g$ then the truncated Taylor expansions of $f\pm g, fg, f/g$ can be found operating on the field $\mathcal P_n$ simply computing $T_f\pm T_g, T_f\cdot T_g, T_f/T_g$. We may thus compute high order derivatives of multivariate rational functions by computing their Taylor expansions in $\mathcal P_n$ and then extracting the desired coefficients.

\subsubsection{Example - A division}
Consider the following rational function of two variables:
$$
h = (x - y) / (x + 2xy + y^2) = f / g
$$
Its Taylor expansion $T_h \in \mathcal P_2$ around the point $x=0$, $y=1$ can be computed as follows:
$$
T_x = 0 + dx
$$
$$
T_y = 1 + dy
$$
$$
T_g =  T_x + 2T_x\cdot T_y + T_y\cdot T_y = 1 + 3 dx + 2 dy +2 dxdy + dy^2 
$$
applying now Eq.(\ref{eq:div}), we get:
$$
T_{1/g} = ( 1 - \hat p + \hat p ^ 2 )
$$
where $\hat p = 3 dx + 2 dy +2 dxdy + dy^2 $, hence:
$$
T_{1/g} = 1 - 3 dx -2 dy + 10dxdy + 9dx^2 + 3dy^2
$$
and,
$$
T_h = (-1+dx-dy) \cdot T_{1/g} = -1+4dx+ dy-9dxdy-12dx^2-dy^2
$$
which allows to compute the value of the function and all derivatives up to the second order in the point $x=0, y=1$:
$$
h = -1, 
\partial_x h = 4,
\partial_y h = 1,
\partial_{xy} h = -9,
\partial_{xx} h = -24,
\partial_{yy} h = -2,
$$
\subsection{Non rational functions}
If the function $f$ is not rational, \emph{i.e.} it is not the ratio between two polynomials, it is still possible, in most cases, to compute its truncated Taylor expansion operating in $\mathcal P_n$. This remarkable result is made possible leveraging on the nil-potency property of $\hat p$ in $\mathcal P_n$, \emph{i.e.} $\hat p^k = 0$ if $k > n$. We outline the derivation in the case of the function $f = \exp(g)$, other cases are treated in details in \cite{textbook}. We write $T_f = \exp(T_g) = \exp(p_0 + \hat p) = \exp p_0 \exp \hat p$. Using the series representation of the function $\exp$ we may write $f = \exp p_0 \sum_{i=0}^\infty \frac{\hat p^i}{i!}$. As per the nil-potency property this infinite series is, in $\mathcal P_n$ finite and we can thus write $\exp(p) = \exp p_0 \sum_{i=0}^n \frac{\hat p^i}{i!}$, or equivalently:
$$
T_f = \exp p_0 \cdot (1 + \hat p + \hat p^2 + .. + \hat p^n)
$$
With similar derivations it is possible to define most commonly used functions including $\exp, \log, \sin, \cos, \tan, \arctan, \arcsin, \arccos$, abs as well as exponentiation. Operating in this algebra, rather than in the common arithmetical one, we compute the programs encoded by CGP obtaining not only the program output, but also all of its differential variations (up to order $n$) with respect to any of the parameters we are interested in. This idea is leveraged in this paper to propose new learning methods in genetic programming.

\section{Computer Implementation}
\label{sec:implementation}
The computer implementation of a CGP system computing in the truncated Taylor polynomial algebra (a dCGP) follows naturally introducing of a new data type representing a truncated polynomial and by overloading all arithmetic operations and functions. Such type is here called \emph{generalized dual number} to explicitly indicate the link with the \emph{dual numbers} often used in computer science to implement first order automatic differentiation. We implemented generalized dual numbers in C++ and Python in the open source project AuDi \cite{audi} and a version of CGP able to compute with them in the open source software dCGP \cite{dCGP}. Besides the basic four arithmetic operations $+,-,\cdot,/$ the following function are, at the moment of writing, available: exp, log, pow, sin, cos, tan, atan, acos, asin, sinh, cosh, tanh, atanh, acosh, asinh, erf, sqrt, cbrt, abs. Two commercial codes are also available that implement the algebra of truncated Taylor polynomials. They were developed for use in beam physics \cite{makino2006cosy} and in astrodynamics \cite{rasotto} and can be compared to our own implementation in AuDi, developed for machine learning applications. When implementing such an algebra, the most sensitive choices are to be made in the implementation of the truncated multiplication between polynomials. Assuming such an operator available, the rest of the algebra is implemented by computing the power series defining the various operators (for example the division would be implemented directly from Eq.(\ref{eq:div})). With respect to the mentioned codes our implementation, differs on the implementation of the truncated multiplication, introducing an approach that allows to parallelize and vectorize the operation. The use of vectorization is particularly useful in the context of machine learning as algorithms often operate on batches to construct a learning error which is typically obtained summing the result of the same polynomial computation repeated over multiple input points.

\begin{table}[t]
\centering
\caption{Speedup obtained by our implementation of the truncated Taylor polynomial differential algebra (AuDi) when computing $\epsilon$ on batches of different sizes $b$, with respect to the library DACE \cite{rasotto}. For small batch sizes and low orders DACE is still two times faster than AuDi.}.
\label{tab:vect_perf}
\begin{tabular}{rc |rrrrrrrrr}
 &order&1 &  2& 3&  4& 5& 6& 7& 8& 9  \\ 
 batch size& \\  \hline
16&&$\mathbf{0.59}$& $\mathbf{0.61}$& $\mathbf{0.65}$& $\mathbf{0.60}$& 1.41& 1.52& 1.86& 1.61& 2.03\\

64&&2.26& 2.07& 2.07& 1.91& 1.51& 2.13& 2.91& 3.04& 3.58\\

256&&5.74& 5.39& 4.82& 4.44& 3.01& 3.64& 4.75& 4.64& 7.18\\

1024&&14.21& 10.82& 8.94& 7.45& 6.70& 6.86& 7.80& 7.52& 8.27\\

4096&&20.02& 13.26& 9.77& 7.44& 4.58& 5.92& 6.15& 5.14& 4.76\\

16384&&19.03& 8.50& 5.34& 4.00& 3.74& 4.71& 4.39& 4.11& 5.01\\
\end{tabular}
\end{table}

\subsection{The truncated multiplication}
Consider the Taylor expansion of a function on $m$ variables in $\mathcal P_n$. The
number $D$ of terms in such a Taylor polynomial is the total number of derivatives of
maximum order $m$ in $n$ variables. Such number is given by the formula:
$$
D = \sum_{k=1}^m {\binom{k+n-1}{k}}
$$
In Table \ref{tav:nmonomials} we report $D$ for varying $n$ and
$m$. It is clear that the explosion of the number of monomials will make the
computations unpractical for high values of $n$ and $m$. Note how all of the existing machine learning methods exploit the upper part of the table, where $m$ is large and, necessarily, $n$ is low (typically maximum two). While the area on Table \ref{tav:nmonomials} that corresponds to low values of $m$ and high orders is unexplored even if the resulting Taylor expansion carries the same number of terms, \emph{i.e.} results in a similar computational complexity.

The details on how to
efficiently implement the truncated polynomial multiplication go beyond the scope of
this paper, the final algorithm here used is made available as part of the open source
algebraic manipulator Piranha \cite{piranha2016}. It is, essentially, based on the work
on algebraic manipulators by Biscani \cite{biscani2010} and exploits the possible
sparsity of the polynomials as well as fine-grained parallelism to alleviate the
intense CPU load needed at high orders/many variables. For the purpose of using Piranha
with AuDi, we added the possibility to compute the truncated multiplication over
polynomials whose coefficients are vectorized, obtaining a substantial advantage in
terms of computational speed as the overhead introduced to order all
monomials in the final expression can be accounted only once per vectorized operation.

\subsubsection{Performances}
In order to preliminarily assess the performance of our implementation of the differential algebra of the truncated Taylor polynomials in AuDi, we measure the CPU time taken to compute the following fictitious training error:
$$
\epsilon = (c + w_1 + w_2 + w_3 + w_4 + w_5 + ..+ w_n)^{N}
$$
which is here taken as representative of the algebraic manipulations necessary to
compute the training error of a program (or model) with $n$ different weights to be
learned. We repeat the same computation on a batch of size $b$ using
DACE\footnote{While a real comparison does not exist in the literature, informal
communications with the authors of DACE revealed that the other existing implementation
of the Taylor polynomials differential algebra, COSY \cite{makino2006cosy} results in
comparable CPU times to DACE} \cite{rasotto} and the AuDi for $N=20$ and $n=7$. 
The truncation order $m$ as well as the batch size $b$ are instead varied. The results
are shown in Table \ref{tab:vect_perf} where it is highlighted how for low batch sizes
and derivation order AuDi does not offer any speedup and is, instead, slower. As soon
as larger batch sizes are needed (already at $b=64$) or higher orders are desired, AuDi
takes instead full advantage of the underlying vectorized truncated polynomial
multiplication algorithm and offers substantial speedup opportunities. The table
reported was obtained on a 40 processors machine equipped with two Intel(R) Xeon(R) CPU
E5-2687W v3 @ 3.10GHz. Different architectures or error definitions may change the
speedup values also considerably, but not the conclusion that at high orders or batch
sizes AuDi is a very competitive implementation of the truncated Taylor polynomials
differential algebra.

\begin{table}[t]
\center
\scriptsize
\begin{tabular}{ll|rrrrrrrrrrrrrr}
\multicolumn{13}{c}{$m$ = number of variables} \\
&& 1  & 2  & 3  & 4  & 5  & 6  & 7  & 8  & 9  & 10  & 11  & 12  & \\ \hline
\multirow{13}{*}{\rotatebox[]{90}{$n$ = order}}\\
&1 & 1  & 2  & 3  & 4  & 5  & 6  & 7  & 8  & 9  & 10  & 11  & 12  & \\
&2&2  & 5  & 9  & 14  & 20  & 27  & 35  & 44  & 54  & 65  & 77  & 90  & \\
&3&3  & 9  & 19  & 34  & 55  & 83  & 119  & 164  & 219  & 285  & 363  & 454  & \\
&4&4  & 14  & 34  & 69  & 125  & 209  & 329  & 494  & 714  & 1000  & 1364  & 1819  & \\
&5&5  & 20  & 55  & 125  & 251  & 461  & 791  & 1286  & 2001  & 3002  & 4367  & 6187  & \\
&6&6  & 27  & 83  & 209  & 461  & 923  & 1715  & 3002  & 5004  & 8007  & 12375  & 18563  & \\
&7&7  & 35  & 119  & 329  & 791  & 1715  & 3431  & 6434  & 11439  & 19447  & 31823  & 50387  & \\
&8&8  & 44  & 164  & 494  & 1286  & 3002  & 6434  & 12869  & 24309  & 43757  & 75581  & 125969  & \\
&9&9  & 54  & 219  & 714  & 2001  & 5004  & 11439  & 24309  & 48619  & 92377  & 167959  & 293929  & \\
&10&10  & 65  & 285  & 1000  & 3002  & 8007  & 19447  & 43757  & 92377  & 184755  & 352715  & 646645  & \\
&11&11  & 77  & 363  & 1364  & 4367  & 12375  & 31823  & 75581  & 167959  & 352715  & 705431  & 1352077  & \\
&12&12  & 90  & 454  & 1819  & 6187  & 18563  & 50387  & 125969  & 293929  & 646645  & 1352077  & 2704155  & \\ \hline
\end{tabular}
\vskip.5cm
\caption{Number of monomials in a Taylor polynomial for varying order $n$ and number of variables $m$. \label{tav:nmonomials}}
\end{table}

\begin{figure}[t]
\centering
      \includegraphics[width=1\textwidth]{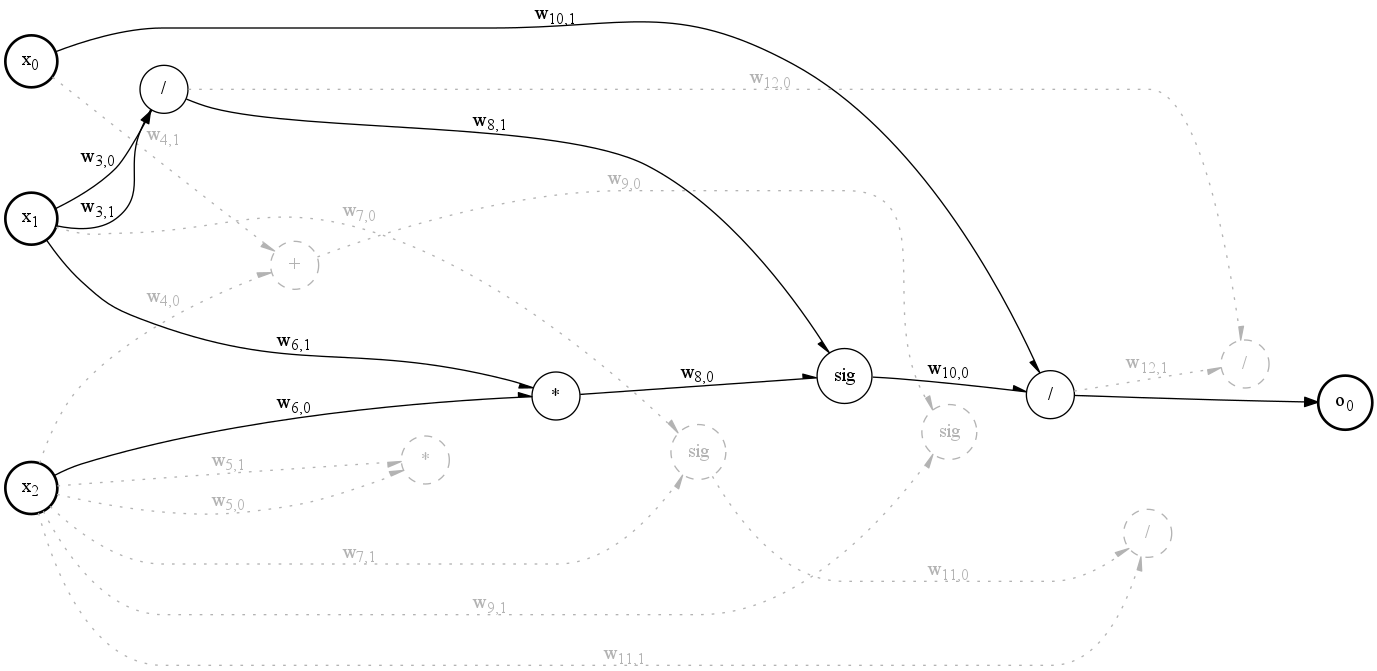} 
      \caption{A weighted CGP graph expressing, (assuming  $w_{i,j}=1$) $o_0 = \frac{\sigma(yz + 1)}{x}$ at the output node.}
      \label{fig:cgp_graph} 
\end{figure}

\section{Example of a dCGP}
\label{sec:example}
Let us consider a CGP expression having $n=3$, $m=1$, $r=1$, $c=10$, $l=3$, $a=2$, $w_{i,j} = 1$ and the following kernel functions: +,*,/,$\sigma$ where $\sigma$ is the sigmoid function. Using the same convention used in \cite{millerbook} for the node numbering, the chromosome:
$$
x = [2, 1, 1, 0, 2, 0, 1, 2, 2, 1, 2, 1, 3, 1, 2, 3, 6, 3, 3, 4, 2, 2, 8, 0, 2, 7, 2, 2, 3, 10, 10]
$$
will produce, after simplifications, the following expression at the terminal node:
$$
o_0 =   \frac{\sigma(yz + 1)}{x}
$$
in Figure \ref{fig:cgp_graph} the actual graph expressing the above expression is visualized. If we now consider the point $x=1$, $y=1$ and $z=1$ and we evaluate classically the CGP expression (thus operating on float numbers), we get, trivially, $O_1 = 0.881$ where, for convenience, we only report the output rounded up to 3 significant digits. Let us, instead, use dCGP to perform such evaluation. One option could be to define $x,y,z$ as generalized dual numbers operating in $\mathcal P_2$. In this case, the output of the dCGP program will then be a Taylor polynomial in $x,y,z$ truncated at second order:
\begin{multline*}
o_0 = 0.881-0.881dx+0.105dy+0.105dz+0.881dx^2 \\ -0.0400dy^2-0.0400dz^2-0.105dxdz+0.025dydz-0.105dxdy
\end{multline*}
carrying information not only on the actual program output, but also on all of its derivatives (in this case up to order 2) with respect to the chosen parameters (\emph{i.e.} all inputs, in this case). 
Another option could be to define some of the weights as generalized dual numbers, in which case it is convenient to report the expression at the output node with the weights values explicitly appearing as parameters:
$$
o_0 = w_{10,0}\frac{\sigma(w_{3,0}w_{8,1}/w_{3,1} + w_{6,0}w_{6,1}w_{8,0}yz)}{w_{10,1}x}
$$
If we define $w_{3,1}$ and $w_{10,1}$ as generalized dual numbers operating in $\mathcal P_3$ then, evaluating the dCGP will result in:
\begin{multline*}
o_0 = 0.881 -0.104dw_{3,1}-0.881dw_{10,1}+0.104dw_{10,1}dw_{3,1}+0.0650dw_{3,1}^2+0.881dw_{10,1}^2 \\ -0.0315dw_{3,1}^3-0.881dw_{10,1}^3-0.104dw_{10,1}^2dw_{3,1}-0.0650dw_{10,1}dw_{3,1}^2
\end{multline*}

\section{Learning constants in symbolic regression}
\label{sec:symbolic}
A first application of dCGP we study is the use of the derivatives of the expressed
program to learn the values of constants by, essentially, back-propagating the error to the constants' value. In 1997, during a
tutorial on Genetic Programming in Paolo Alto, John Koza \cite{koza1997tutorial} one of the fathers of Genetic Programming
research noted that \lq\lq finding of numeric constants is a skeleton in the GP closet ... an area of research that requires
more investigation\rq\rq. Years later, the problem, while investigated by multiple researchers, is still
open \cite{o2010open}. The standard approach to this issue is that of the \emph{ephemeral constant}
where a few additional input terminals containing constants are added and used by the encoded program to
build more complex representations of any constant. Under this approach one would hope that to
approximate, say, $\pi$ evolution would assemble, for example, the block $\frac{22}7 = 3.1428$ from the additional
terminals. In some other versions of the approach the values of the input constants are subject to
genetic operators \cite{erc2}, or are just random. Here we first present a new approach that back-propagates the error on the ephemeral constants values, later we introduce a different approach using weighted dCGP expressions and back-propagating the error on the weights. 

\subsection{Ephemeral constants approach}
The idea of learning ephemeral constants by back-propagating the
error of a GP expression was first studied in \cite{erc1} where gradient descent was used during evolution of a GP tree. The technique was not proved, though,
to be able to solve the problems there considered (involving integers), but to reduce the RMSE at a fixed
number of generations with respect to a standard genetic programming technique. 
Here we will, instead, focus on using dCGP to solve exactly symbolic regression problems involving real
constants such as $\pi$ and $e$.
Consider the quadratic error $\epsilon_q (\mathbf c) = \sum_i (y_i(\mathbf c) - \hat y_i)^2$
where $\mathbf c$ contains the values of the ephemeral constants, $y_i$ is the value of the
dCGP evaluated on the input point $x_i$ and $\hat y_i$ the target value. Define the fitness (error) of a candidate
dCGP expression as:
$$
\epsilon = \min_{\mathbf c} \epsilon_q(\mathbf c)
$$
This "inner" minimization problem can be efficiently solved by a second order method.
If the number of input constants is reasonably small, the classic formula
for the Newton's method can be conveniently applied iteratively:
$$
\mathbf c_{i+1} = \mathbf c_i - \mathbf H^{-1}(\mathbf c_i) \nabla \epsilon_q (\mathbf c_i)
$$
starting from some initial value $\mathbf c_0$. The Hessian $\mathbf H$ and the gradient $\nabla \epsilon$
are extracted from the Taylor expansion of the error $\epsilon$ computed via
the dCGP. Note that if the constants $\mathbf c$ appear only linearly in the candidate expression, one
single step will be sufficient to get the exact solution to the inner minimization problem, as 
$\epsilon_q(c)$ is the quadratic error.
This suggests the use of the following approximation of the error defined above, where one single
learning step is taken:
\begin{equation}
\label{eq:qe_erc}
\epsilon = \epsilon_q(\mathbf c_0 - \mathbf H^{-1}(\mathbf c_0) \nabla \epsilon_q (\mathbf c_0))
\end{equation}
where $\mathbf c_0$ is the current value for the constants.
In those cases where this Newton step does not reduce the error (when $\mathbf H$ is not
positive definite we do not have any guarantee that the search direction will lead to a
smaller error) we, instead, take a few steps of gradient descent (learning rate set to 0.05).
The value $\mathbf c_0$ is initialized to some random value and, during evolution, is set to
be the best found so far. Note that when one offspring happen to be the un-mutated
parent, its fitness will still improve on the parents' thanks to the local learning, essentially by applying one or 
more Newton step. This mechanism (also referred to as Lamarckian evolution in memetic
research \cite{erc1}) ensures that the exact value, and not an approximation, is
eventually found for the constants even if only one Newton step is taken at each iteration. Note also
that the choice of the quadratic error function in connection with 
a Newton method will greatly favour, during evolution, expressions such as, for example, $c_1 x + c_2$
rather than the equivalent $c_1^2 c_2 x + \frac 1{c_2}$.

\begin{table}[t]
\caption{Learning the ephemeral constants: experiments definition and results. In all cases the constants and the
expressions are found exactly
(final error is $\epsilon < 10^{-14}$). \label{tab:erc}}
\begin{tabular}{p{0.5cm}|p{2.5cm}|p{2.5cm}|p{3cm}|c|c|c}
 & target  & expression found & constants
 found & ERT & bounds & $g_{max}$\\ \hline
P1: & $x^5 - \pi x^3 + x$  & $-cx^3 + x^5 + x$ & $c=3.1415926$ & 19902 & [1,3] & 1000\\
P2: & $x^5 - \pi x^3 + \frac{2\pi}x$ & $cx^3 - 2c/x + x^5$&  $c=-3.1415926$ & 124776 & [0.1,5] & 5000 \\
P3: & $\frac{e x^5 + x^3}{x + 1}$ & $\frac{-c x^5 + x^3}{x + 1}$ & $c=-2.7182818$ & 279400 & [-0.9,1] & 5000\\
P4: &  $\sin(\pi x) + \frac 1x$ & $\sin(cx + x) + \frac 1x$ & $c=2.1415926$ & 105233 &[-1,1] & 5000 \\ \hline
P5: & $e x^5 - \pi x^3 + x$ & $c_1x^3 - c_2x^5 + x$ &  $c_1 = -3.1415926$  & 45143  & [1,3] & 2000 \\
&&&$c_2= -2.7182818$&&&\\
P6: & $\frac{e x^2-1}{\pi (x + 2)}$ & $\frac{c_1+c_2 x}{x + 2}$ & $c_1 = -0.3183098$  & 78193 & $[-2.1,1]$ & 10000\\
&&&$c_2 = 0.86525597$&&&\\ 
P7: & $\cos(\pi x) + \sin(e x)$ & $\sin(c_1^2c_2x + c_1^2x)$ & $c_1 = 1.7724538$& 210123 & [-1,1] & 5000\\
&& $+ \cos(c_1^2x)$ & $c_2=-0.1347440$ &&& \\\hline
\end{tabular}
\end{table}

\subsubsection{Experiments} We consider seven different symbolic regression problems (see Table
\ref{tab:erc}) of varying complexity and containing the real constants $\pi$ and $e$. Ten points $x_i$
are chosen on a uniform grid within some lower and upper bound. For all problems the error defined by
Eq.(\ref{eq:qe_erc}) is used as fitness to evolve an unweighted dCGP
with $r=1$, $c=15$, $l=16$, $a=2$, $N_{in} = 2$ or $3$ according to the number of
constants present in the target expression and $N_{out}=1$. As
Kernel functions we use: $+,-,*,/$ for problems P1, P2, P3, P5, P6 and $+,-,*,/,\sin,\cos$ for problems
P4, P7. The evolution is driven by a (1+$\lambda$)-ES evolution strategy with $\lambda=4$ where
the $i-th$ mutant of the $\lambda$ offspring is created by mutating $i$ active genes in
the dCGP. We run all our experiments\footnote{The exact details on
all these experiments can be found in two IPython notebooks available here: \url{https://goo.gl/iH5GAR}, \url{https://goo.gl/0TFsSv}}
100 times and for a maximum of $g_{max}$ generations. The 100 runs are then considered as multi starts of
the same algorithm and are used to compute the Expected Run Time \cite{auger2005performance} (ERT),
that is the ratio between the overall number of dCGP evaluations made ($fevals$) made across
all trials and the number of successful trials $n_s$: $ERT = \frac{fevals}{n_{s}}$. As shown in Table \ref{eq:qe_erc} our approach is able to solve all problems exactly (final error is $\epsilon < 10^{-14}$) and to represent the real constants with precision.

\subsection{Weighted dCGP approach}
While the approach we developed above works very well for the selected
problems, it does have a major drawback: the number of ephemeral constants to be used as
additional terminals must be pre-determined. If one were to use too few ephemeral constants
the regression task would fail to find a zero error, while if one were to put too many
ephemeral constants the complexity would scale up considerably and the proposed solution
strategy would quickly lose its efficacy. This is particularly clear in the comparison
between problems P1 and P5 where the only difference is the use of a multiplying factor $e$
in front of the quintic term of the polynomial. Ideally this should not change the
learning algorithm. As detailed in Sections \ref{sec:encoding} and \ref{sec:example}, a dCGP
gives the possibility to associate weights to each edge of the acyclic graph (see Figure
\ref{fig:cgp_graph}) and then compute the differential properties of the error w.r.t. any
selection of the weights.  This introduces the idea of performing symbolic regression tasks
using a weighted dCGP expression with no additional terminal inputs but with the additional
problem of having to learn values for all the weights appearing in the represented
expression. In particular we now have to
define the fitness (error) of a candidate dCGP expression as:
$$
\epsilon = \min_{\mathbf w} \epsilon_q(\mathbf w)
$$
where $\mathbf w$ are the weights that appear in the output terminal. Similarly to what we did previously, we need a way to solve this inner minimization problem. Applying a few Newton steps could be an option, but since the number of weights may be large we will not follow this idea. Instead, we iteratively select, randomly, $n_w$ weights $\mathbf{\tilde w}$ active in the current dCGP expression and we update them applying one Newton step:
$$
\mathbf{\tilde w_{i+1}} = \mathbf{\tilde w_{i}} - \mathbf{\tilde H}^{-1}(\mathbf{\tilde w_{i}}) \nabla_{ \mathbf{\tilde w}} \epsilon_q (\mathbf{\tilde w_{i}})
$$
where the tilde indicates that not all, but only part of the weights are selected. If no improvement is found we discard
the step. At each iteration we select randomly new weights (no Lamarckian learning). This idea (that we call weight batch learning), is effective in our case as it avoids computing and inverting large Hessians, while also having more chances to actually
improve the error at each step without the use of a learning rate for a line search. For each candidate expression we perform $N$ iterations of weight batch learning starting from normally distributed initial values for the weights.

\begin{table}[t]
\caption{Learning constants using the weighted dCGP: experiments definition and results. In all cases
the constants and the expressions are found exactly (final error is $\epsilon < 10^{-14}$)
\label{tab:wdcgp}}
\begin{tabular}{p{0.5cm}|p{2.5cm}|p{3.3cm}|p{2.38cm}|c|c|c}
 & target & expression found & constants found & ERT & bounds & $g_{max}$\\ \hline
P1: & $x^5 - \pi x^3 + x$  & $c_1x^5 + c_2x^3 + c_3x$ & $c_1=1.0$ & 106643 & [1,3] & 200\\
&&& $c_2=-3.1415926$ &&&\\
&&& $c_3=0.9999999$ &&&\\
P2: & $x^5 - \pi x^3 + \frac{2\pi}x$ & $c_1x^5 + c_2x^3 + \frac{c_3}{c_4 x}$ & $c_1=0.9999999$ & 271846 & [0.1,5] & 200 \\
&&& $c_2=-3.1415926$ &&&\\
&&& $c_3=6.2831853$ &&&\\
&&& $c_4=1.0$ &&&\\
P3: & $\frac{e x^5 + x^3}{x + 1}$ & $\frac{c_1x^5 + c_2x^3}{c_3x + c_4}$ & $c_1=4.3746892$ & 1935500 & [-0.9,1] & 200\\
&&& $c_2=1.6093582$ &&&\\
&&& $c_3=1.6093582$ &&&\\
&&& $c_4=1.6093582$ &&&\\
P4: &  $\sin(\pi x) + \frac 1x$ & $c_1 \sin(c_2x) + \frac{c_3}{c_4 x}$ & $c_1=1.0$ & 135107 &[-1,1] & 200 \\
&&& $c_2=3.1415926$ &&&\\
&&& $c_3=1.0$ &&&\\
&&& $c_4=1.0$ &&&\\
P5: & $e x^5 - \pi x^3 + x$ & $c_1x^5 + c_2x^3 + c_3x$ & $c_1=2.7182818$ & 122071 & [1,3] & 200 \\
&&& $c_2=-3.1415926$ &&&\\
&&& $c_3=1.0$ &&&\\
P6: & $\frac{e x^2-1}{\pi (x + 2)}$ & $\frac{c_1x^2 + c_2}{c_3x + c_4}$ & $c_1=1.5963630$ & 628433 & [-2.1,1] & 200\\
&&& $c_2=-0.5872691$ &&&\\
&&& $c_3=1.8449604$ &&&\\
&&& $c_4=3.6899209$ &&&\\
P7: & $\cos(\pi x) + \sin(e x)$ & $c_1\sin(c_2x) + c_3\cos(c_4x)$ & $c_1=1.0$ & 243629 & [-1,1] & 200\\
&&& $c_2=2.7182818$ &&&\\
&&& $c_3=1.0$ &&&\\
&&& $c_4=3.1415926$ &&&\\ \hline
\end{tabular}
\end{table}

\subsubsection{Experiments} 
We use the same experimental set-up employed to perform symbolic regression using the ephemeral
constants approach (see Table \ref{tab:wdcgp}). For each iteration of the weight batch learning we use a
randomly selected $n_w \in \{2, 3\}$ and we perform $N = 100$ iterations\footnote{The exact details on
all these experiments can be found in the IPython notebook available here: \url{https://goo.gl/8fOzYM}}. 
The initial weights are drawn from a
zero mean, unit standard deviation normal distribution. By assigning weights to all the edges, we end up with expressions where every
term has its own constant (\emph{i.e.}, a combination of weights), hence no distinction is made by the learning algorithm between integer and real constants. This gives the method a far greater generality than
the ephemeral constants approach as the number of real constants appearing in the target
expression is not used as an information to design the encoding. It is thus not
a surprise that we obtain, overall, higher ERT values. These higher values mainly
come from the Newton steps applied on each weighted candidate expression and not from the
number of required generations which was, instead, observed to be generally much lower
across all experiments. Also note that for problems P3 and P6 the final expression found can
have an infinite number of correct constants, obtained by multiplying the numerator and
denominator by the same number. Overall, the new method here proposed to perform symbolic
regression on expressions containing constants was able to consistently solve the
problems considered and is also suitable for applications where the number of real constants to be learned is not known in advance.

\section{Solution to Differential Equations}
\label{sec:de}
We show how to use dCGP to search for expressions $S(x_1, .., x_n) = S(\mathbf x)$ that solve a generic differential equation of order $m$ in the form:
\begin{equation}
\label{eq:probdef}
f\left(\partial^\alpha S, \mathbf x \right) = 0, |\alpha| \le m
\end{equation}
with some boundary conditions $S(\mathbf x) = S_{\mathbf x}, \forall \mathbf x \in \mathcal B$. Note that we made use of the multi-index notation for high order
derivatives described previously. Such formal representation includes initial value
problems for ordinary differential equations and boundary value problems for partial
differential equations. While this representation covers only the case of Dirichelet
boundary conditions (values of $S$ are specified on a border), the system devised
here can be used also for Neumann boundary conditions (values of $\partial S$ are
specified on a border). Similarly, also systems of equations can be studied. Assume $S$ is encoded by a dCGP expression: one may easily compute Eq.(\ref{eq:probdef})
over a number of control points placed in some domain, and boundary values $S(\mathbf x)$ may also be computed on some other control points placed over $\mathcal B$. It is thus natural to compute the following expression and use it as error: 
\begin{equation}
\label{eq:errorDE}
\epsilon = \sum_i f^2\left(\partial^\alpha S, \mathbf x_i \right) + \alpha \sum_j \left(S(\mathbf x)-S_{\mathbf x}\right)^2
\end{equation}
which is, essentially, the sum of the quadratic error of the violation of the differential equations plus the quadratic error of the violation on the boundary values. Symbolic regression was studied already in the past as a new tool to find solutions to differential equations by Tsoulos and Lagaris \cite{tsoulos} who used grammatical
evolution to successfully find solutions to a large number of ordinary differential
equations (ODEs and NLODEs), partial differential equations (PDE), and systems of
ordinary differential equations (SODEs). To find the derivatives of the encoded expression
Tsoulos and Lagaris add additional stacks where basic rules for 
differentiation are applied in a chain. Note that such system (equivalent to a basic automatic
differentiation system) is not easily extended to the computation of mixed
derivatives, necessary for example when Neumann boundary conditions are present. As a consequence, all of the differential test problems introduced in
\cite{tsoulos} do not involve mixed derivatives. To test the use of dCGP on this
domain, we consider a few of those problems and compare our results to the ones
obtained by Tsoulos and Lagaris. From the paper it is possible to derive the average
number of evaluations that were necessary to find a solution to the given problem, by
multiplying the population size used and the average number of generations reported. This
number can be then compared to the ERT \cite{auger2005performance} obtained in our multi-start experiments.

\subsubsection{Experiments}
For all studied problems we use the error defined by Eq.(\ref{eq:errorDE})
to train an unweighted dCGP with $r=1$, $c=15$,
$l=16$, $a=2$, $N_{in}$ equal to the number of input variables and $N_{out}=1$. As
Kernel functions we use the same as that used by Tsoulos and Lagaris: $+,-,*,/,\sin,\cos,\log,\exp$. The
evolution is made using a (1+$\lambda$)-ES evolution strategy with $\lambda=10$ where
the $i-th$ mutant of the $\lambda$ offspring is created mutating $i$ active genes in
the dCGP. We run all our experiment\footnote{The full details on
these experiments can be found in an IPython notebook available here: \url{https://goo.gl/wnCkO9}}
100 times and for a maximum of 2000 generation.
We then record the successful runs (\emph{i.e.} the runs in which the error is reduced
to $\epsilon \le 10^{-16}$) and compute the expected run-time as the ratio
between the overall number of evaluation of the encoded program and its derivatives ($fevals$) made across successful and unsuccessful trials and the number of successful trials $n_s$: $ERT = \frac{fevals}{n_{s}}$ \cite{auger2005performance}. The
results are shown in Table \ref{tab:ERTde}. It is clear how, in all test
cases, the dCGP based search is able to find solutions very efficiently, outperforming the baseline results.

\begin{table}[t]
\centering
\caption{Expected run-time for different test cases taken from \cite{tsoulos}. The ERT can be seen as the average number of evaluation of the program output and its derivatives needed (on average) to reduce the error to zero (\emph{i.e.} to find an exact solution)}
\label{tab:ERTde}
\begin{tabular}{l|c|c}
 Problem & d-CGP & Tsoulos \cite{tsoulos}  \\ \hline
ODE1 & 8123 & 130600   \\
ODE2 & 35482 & 148400   \\
ODE5 & 22600 & 88200 \\
NLODE3 & 896 & 38200 \\
PDE2 & 24192 & 40600 \\
PDE6 & 327020 & 797000 \\ \hline
\end{tabular}
\end{table}

\section{Discovery of prime integrals}
\label{sec:primeintegrals}
We now show how to use dCGP to search for expressions $P$ that are prime integrals of some set of differential equations. Consider a set of ordinary differential equations (ODEs) in the form:
$$\left\{
\begin{array}{c}
\frac{dx_1}{dt} = f_1(x_1, \cdots, x_n) \\
\vdots \\
\frac{dx_n}{dt} = f_n(x_1, \cdots, x_n)
\end{array}\right.
$$
Solutions to the above equations are denoted with $x_i(t)$ to highlight their time dependence. Under relatively loose conditions on the functions $f_i$, the solution to the above equations always exists unique if initial conditions $x_i(t_0)$ are specified. A prime integral for a set of ODEs is a function of its solutions in the form $P(x_1(t), \cdots, x_n(t)) = 0, \forall t$. Prime integrals, or integral of motions, often express a physical law such as energy or momentum conservation, but also the conservation of some more \lq\lq hidden\rq\rq quantity as its the case in the Kovalevskaya top \cite{borisov2001sv} or of the Atwood's machine \cite{casasayas1990swinging}. Each of them allows to decrease the number of degrees of freedom in a system by one. In general, finding prime integrals is a very difficult task and it is done by skillful mathematicians using their intuition. A prime integral is an implicit relation between the variables $x_i$ and, as discussed by Schmidt and Lipson \cite{schmidt2010symbolic}, when symbolic regression is asked to find such implicit relations, the problem of driving evolution towards non trivial, informative solutions arises. We thus have to devise an experimental set-up that is able to avoid such a behaviour.
Assuming $P$ to be a prime integral, we differentiate it obtaining:
$$
\frac{dP}{dt} = \sum_i \frac{\partial P}{\partial x_i}
 \frac {dx_i}{dt} = \sum_i \frac{\partial P}{\partial x_i} f_i = 0
$$
The above relation, and equivalent forms, is essentially what we use to compute the error of some candidate expression representing a prime integral $P$. In more details, we define $N$ points in the phase space and denote  them with $x_i^j, j=1..N$. Thus, we introduce the error function:
\begin{equation}
\label{eq:error}
\epsilon = \sum_j \sum_i \left[\frac{\partial P}{\partial x_i}(x_1^j, ..., x_n^j) f_i(x_1^j, ..., x_n^j)\right]^2
\end{equation}
Since the above expression is identically zero if $P$ is, trivially, a constant we introduce a \lq\lq mutation suppression\rq\rq\ method during evolution. Every time a new mutant is created, we compute  $\frac {\partial P}{\partial x_i}, \forall i$ and we ignore it if $\frac {\partial P}{\partial x_i} = 0, \forall i$, that is if the symbolic expression is representing a numerical constant. Our method bears some resemblance to the approach described in \cite{lipson} where physical laws are searched for to fit some observed system, but it departs in several significant points: we do not use experimental data, rather the differential description of a system, we compute our derivatives using the Taylor polynomial algebra (\emph{i.e.} automatic differentiation) rather than estimating them numerically from observed data, we use the mutation suppression method to avoid evolving trivial solutions, we need not to introduce a \emph{variable pairing} \cite{lipson} choice and we do not assume variables as not dependent on all others. All the experiments\footnote{The full details on our experiments can be found in an IPython notebook available here: \url{https://goo.gl/ATrQR5}}
are made using a
non-weighted dCGP with $N_{out} = 1$, $r=1$, $c=15$, $lb=16$, $a=2$. We use a
(1+$\lambda$)-ES evolution strategy with $\lambda=10$ where the $i-th$ mutant of the
$\lambda$ offspring is created mutating $i$ active genes in the dCGP. A set of $N$
control points are then sampled uniformly in some bounds. Note how these are not
belonging to any actual trajectory of the system, thus we do not need to numerically
integrate the ODEs. We consider three different dynamical systems: a mass-spring
system, a simple pendulum and the two body problem. 

\subsubsection{Mass-spring system}
Consider the following equations:
$$
MSS: \left\{
\begin{array}{l}
\dot v = -kx \\
\dot x = v
\end{array}\right.
$$
describing the motion of a simple, frictionless mass-spring system (MSS). We use $N=50$ and create the points at random as follows: $x \in [2,4]$, $v \in [2,4]$ and $k \in[2, 4]$. For the error, rather than using directly the form in Eq.(\ref{eq:error}) our experiments indicate that a more informative (and yet mathematically equivalent) variant is, in this case, $\epsilon = \sum_j \left[ \frac{ \frac{\partial P}{\partial r}
}{ \frac{\partial P}{\partial v}
} + \frac{f_v}{f_r}\right]^2$. 
\small
\begin{table}[t]
\label{tab:resultspi}
\caption{Results of the search for prime integrals in the mass-spring system (MSS), the simple pendulum (SP), and two body problem (TBP). Some prime integrals found are also reported in both the original and simplified form. The Expected Run Time (ERT), \emph{i.e.} the number of evaluations of the dCGP expression that is, on average, required to find a prime integral, is also reported. }
MSS: Energy Conservation (ERT=50000)\\
\begin{tabular}{p{8cm}|p{4cm}}
Expression found & Simplified \\ \hline
$k*((x*x)-k)+k+(v*v)$  &  $-k^2 + kx^2 + k + v^2$ \\
$(x*x)+(v/k)*v$  &  $x^2 + v^2/k$ \\
$(k*((x*x)/v)/(k*(x*x)/v)+v)/(x*x)$  &  $k/(kx^2 + v^2)$ \\
$(v+x)*((v+x)-x)-x*(((v+x)-x)-(x*k))$  &  $kx^2 + v^2$ 
\end{tabular}\\ \\
SP: Energy Conservation (ERT=114000)\\
\begin{tabular}{p{8cm}|p{4cm}}
Expression found & Simplified \\ \hline
$((((\omega*\omega)/c)-\cos(\theta))-\cos(\theta))/c$ &  $-2\cos(\theta)/c + \omega^2/c^2$ \\
$(\omega/(c+c))*(\omega-(\cos(\theta)/(\omega/(c+c))))$ &  $-\cos(\theta) + \omega^2/c$\\
$((\omega*\omega)-((c+c)*\cos(\theta)))$  & $-2c\cos(\theta) + \omega^2$\\
$(((c+c)*\cos(\theta))-(\omega*\omega))-\sin((\theta/\theta))$ &  $2c\cos(\theta) - \omega^2 - \sin(1)$
\end{tabular}\\ \\
TBP: Angular momentum Conservation (ERT=5270)\\
\begin{tabular}{p{8cm}|p{4cm}}
Expression found & Simplified \\ \hline
$(((\mu/r)/(r/\mu))/\omega)$&$\mu^2/(\omega r^2)$ \\
$((((r*r)*\omega)+\mu)/\mu)$&$1 + \omega r^2/\mu$\\
$((\omega*\mu)*((r*r)*\mu))$&$\mu^2*\omega r^2$\\
$((\mu/(\mu+\mu))-((r*\omega)*r))$&$-\omega r^2 + 1/2$
\end{tabular}\\ \\
TBP: Energy Conservation (ERT=6144450)\\
  \begin{tabular}{p{8cm}|p{4cm}}
Expression found & Simplified \\ \hline
$((v-r)*(v+r))-(((\mu-r)+(\mu-r))/r-((r-(r*\omega))*(r-(r*\omega)))) $ & $-2\mu/r + \omega^2r^2 - 2\omega r^2 + v^2 + 2$\\
$((\mu/r)-((v*v)-(\mu/r)))-((r*\omega)*(r*\omega))$ & $2\mu/r - \omega^2 r^2 - v^2$\\
$(((\omega*r)*(r-(\omega*r)))+((\mu/r)-((v*v)-(\mu/r))))$ & $2\mu/r - \omega^2 r^2 + \omega r^2 - v^2$\\
$(((r*\omega)*\omega)*r)-(((\mu/r)+(\mu/r))-(v*v))$  & $-2\mu/r + \omega^2 r^2 + v^2$
\end{tabular}
\end{table}
\normalsize

\subsubsection{Simple pendulum}
Consider the following equations:
$$SP: \left\{
\begin{array}{l}
\dot \omega = - \frac gL\sin\theta \\
\dot \theta = \omega \\
\end{array}\right.
$$
describing the motion of a simple pendulum (SP). We use $N=50$ and create the points at random as follows: $\theta \in [-5,5]$, $\omega \in [-5,5]$ and $c =\frac gL \in(0, 10]$. Also in this case, we use a variant for the error expression:  $\epsilon = \sum_j \left[ \frac{ \frac{\partial P}{\partial \theta}
}{ \frac{\partial P}{\partial \omega}
} + \frac{f_\omega}{f_\theta}\right]^2$. 

\subsubsection{Two body problem}
Consider the following equations:
$$TBP: \left\{
\begin{array}{l}
\dot v = -\frac\mu{r^2} + r\omega^2 \\
\dot \omega = - 2 \frac{v\omega}{r} \\
\dot r = v \\
\dot \theta = \omega
\end{array}\right.
$$
describing the Newtonian motion of two bodies subject only to their own mutual gravitational interaction (TBP). We use $N=50$ random control of points sampled uniformly as follows: $r \in [0.1,1.1]$, $v \in [2,4]$, $\omega \in [1,2]$ and $\theta \in[2, 4]$ and $\mu \in [1,2]$ (note that these conditions allow for both elliptical and hyperbolic motion). The unmodified form for the error (Eq.(\ref{eq:error})) is used at first and leads to the identification of a first prime integral (the angular momentum conservation). Since the evolution quickly and consistently converges to that expression, the problem arises on how to find possibly different ones. Changing the error expression to $\epsilon = \sum_j \left[ 
\frac{ \frac{\partial P}{\partial r}}{ \frac{\partial P}{\partial v}} 
+ \frac{f_v}{f_r} 
+ \frac{ \frac{\partial P}{\partial \theta}}{ \frac{\partial P}{\partial v}} \frac{f_\theta}{f_r} + \frac{ \frac{\partial P}{\partial \omega}}{ \frac{\partial P}{\partial v}} \frac{f_\omega}{f_r}
\right]^2$ forces evolution away from the angular conservation prime integral and thus allows for other expressions to be found.

\subsubsection{Experiments}
For each of the above problems we run 100 independent experiments letting the dCGP expression
evolve up to a maximum of $2000$ generations (brought up to $100000$ for the most difficult case of
the second prime integral in the two-body problem). We record the successful runs and the
generation number to then evaluate the expected run time (ERT) \cite{auger2005performance}. The
results are shown in Table \ref{tab:resultspi} where we also report some of the expressions found
and their simplified form. In all cases the algorithm is able to find all prime integrals with the
notable case of the two-body problem energy conservation being very demanding in terms of the  ERT.
In this case we note how the angular momentum $\omega r^2$ is often present in the final expression
found as it there acts as a constant. The systematic search for prime integrals is successful in
these cases and makes use of no mathematical insight into the system studied, introducing a new
computer assisted procedure that may help in future studies of dynamical systems.

\section{Conclusions}
We have introduced a novel machine learning framework called differentiable genetic programming, which
makes use of a high order automatic differentiation system to compute any-order derivatives of the
program outputs and errors. We test the use of our framework on three distinct open problems
in symbolic regression: learning constants, solving differential equations and searching for
physical laws. In all cases, we find our model able to successfully solve selected problems and,
when applicable, to outperform previous approaches. Of particular interest is the novel solution proposed to the long debated problem of constant finding in genetic programming, here proved to allow to find the exact solution in a number of interesting cases. Our work
is a first step towards the systematic use of differential information in learning algorithms for
genetic programming.

\bibliographystyle{splncs03}

%\begin{thebibliography}{}
\bibliography{biblio}

\begin{thebibliography}{10}
\providecommand{\url}[1]{\texttt{#1}}
\providecommand{\urlprefix}{URL }

\bibitem{auger2005performance}
Auger, A., Hansen, N.: Performance evaluation of an advanced local search
  evolutionary algorithm. In: 2005 IEEE congress on evolutionary computation.
  vol.~2, pp. 1777--1784. IEEE (2005)

\bibitem{textbook}
Bertz, M.: Modern map methods in particle beam physics, vol. 108. Academic
  Press (1999)

\bibitem{biscani2010}
Biscani, F.: Parallel sparse polynomial multiplication on modern hardware
  architectures. In: Proceedings of the 37th International Symposium on
  Symbolic and Algebraic Computation. pp. 83--90. ISSAC '12, ACM, New York, NY,
  USA (2012), \url{http://doi.acm.org/10.1145/2442829.2442845}

\bibitem{piranha2016}
Biscani, F., Fernando, I., \v{C}ert\'{i}k, O.: piranha: 0.2 (Jun 2016),
  \url{https://doi.org/10.5281/zenodo.56369}

\bibitem{borisov2001sv}
Borisov, A.V., Kholmskaya, A., Mamaev, I.S.: Sv kovalevskaya top and
  generalizations of integrable systems. Regular and Chaotic Dynamics  6(1),
  1--16 (2001)

\bibitem{casasayas1990swinging}
Casasayas, J., Nunes, A., Tufillaro, N.: Swinging atwood's machine:
  integrability and dynamics. Journal de Physique  51(16),  1693--1702 (1990)

\bibitem{erc2}
Cerny, B.M., Nelson, P.C., Zhou, C.: Using differential evolution for symbolic
  regression and numerical constant creation. In: Proceedings of the 10th
  annual conference on Genetic and evolutionary computation. pp. 1195--1202.
  ACM (2008)

\bibitem{graves1}
Graves, A., Wayne, G., Danihelka, I.: Neural turing machines. arXiv preprint
  arXiv:1410.5401  (2014)

\bibitem{graves2}
Graves, A., Wayne, G., Reynolds, M., Harley, T., Danihelka, I.,
  Grabska-Barwi{\'n}ska, A., Colmenarejo, S.G., Grefenstette, E., Ramalho, T.,
  Agapiou, J., et~al.: Hybrid computing using a neural network with dynamic
  external memory. Nature  538(7626),  471--476 (2016)

\bibitem{hansen2006cma}
Hansen, N.: The cma evolution strategy: a comparing review. In: Towards a new
  evolutionary computation, pp. 75--102. Springer (2006)

\bibitem{dCGP}
Izzo, D.: {dCGP}: First release (Nov 2016),
  \url{https://doi.org/10.5281/zenodo.164627}

\bibitem{audi}
Izzo, D., Biscani, F.: {AuDi}: First release (Nov 2016),
  \url{https://doi.org/10.5281/zenodo.164628}

\bibitem{cgpannkahn}
Khan, M.M., Ahmad, A.M., Khan, G.M., Miller, J.F.: Fast learning neural
  networks using cartesian genetic programming. Neurocomputing  121,  274--289
  (2013)

\bibitem{koza1997tutorial}
Koza, J.: Tutorial on advanced genetic programming, at genetic programming
  1997. Palo Alto, CA  (1997)

\bibitem{makino2006cosy}
Makino, K., Berz, M.: Cosy infinity version 9. Nuclear Instruments and Methods
  in Physics Research Section A: Accelerators, Spectrometers, Detectors and
  Associated Equipment  558(1),  346--350 (2006)

\bibitem{millerbook}
Miller, J.F.: Cartesian genetic programming. In: Cartesian Genetic Programming,
  pp. 17--34. Springer (2011)

\bibitem{o2010open}
O’Neill, M., Vanneschi, L., Gustafson, S., Banzhaf, W.: Open issues in
  genetic programming. Genetic Programming and Evolvable Machines  11(3-4),
  339--363 (2010)

\bibitem{rasotto}
Rasotto, M., Morselli, A., Wittig, A., Massari, M., Di~Lizia, P., Armellin, R.,
  Valles, C., Ortega, G.: Differential algebra space toolbox for nonlinear
  uncertainty propagation in space dynamics. In: Proceedings of the 6th
  International Conference on Astrodynamics Tools and Techniques (ICATT),
  Darmstadt (2016)

\bibitem{lipson}
Schmidt, M., Lipson, H.: Distilling free-form natural laws from experimental
  data. science  324(5923),  81--85 (2009)

\bibitem{schmidt2010symbolic}
Schmidt, M., Lipson, H.: Symbolic regression of implicit equations. In: Genetic
  Programming Theory and Practice VII, pp. 73--85. Springer (2010)

\bibitem{erc1}
Topchy, A., Punch, W.F.: Faster genetic programming based on local gradient
  search of numeric leaf values. In: Proceedings of the Genetic and
  Evolutionary Computation Conference (GECCO-2001). vol. 155162 (2001)

\bibitem{tsoulos}
Tsoulos, I.G., Lagaris, I.E.: Solving differential equations with genetic
  programming. Genetic Programming and Evolvable Machines  7(1),  33--54 (2006)

\bibitem{cgpannturner}
Turner, A.J., Miller, J.F.: Cartesian genetic programming encoded artificial
  neural networks: a comparison using three benchmarks. In: Proceedings of the
  15th annual conference on Genetic and evolutionary computation. pp.
  1005--1012. ACM (2013)

\end{thebibliography}

\end{document}